# Training-free Quantum-Inspired Image Edge Extraction Method

Arti Jain, Pradeep Singh, Senior Member, IEEE

*Abstract*—Edge detection is a cornerstone of image processing, yet existing methods often face critical limitations. Traditional deep learning edge detection methods require extensive training datasets and fine-tuning, while classical techniques often fail in complex or noisy scenarios, limiting their real-world applicability. To address these limitations, we propose a training-free, quantum-inspired edge detection model. Our approach integrates classical Sobel edge detection, the Schrödinger wave equation refinement, and a hybrid framework combining Canny and Laplacian operators. By eliminating the need for training, the model is lightweight and adaptable to diverse applications. The Schrödinger wave equation refines gradient-based edge maps through iterative diffusion, significantly enhancing edge precision. The hybrid framework further strengthens the model by synergistically combining local and global features, ensuring robustness even under challenging conditions. Extensive evaluations on datasets like BIPED, Multicue, and NYUD demonstrate superior performance of the proposed model, achieving state-of-the-art metrics, including ODS, OIS, AP, and F-measure. Noise robustness experiments highlight its reliability, showcasing its practicality for real-world scenarios. Due to its versatile and adaptable nature, our model is well-suited for applications such as medical imaging, autonomous systems, and environmental monitoring, setting a new benchmark for edge detection.

## I. INTRODUCTION

EDGE extraction plays a pivotal role in computer vision and image processing, serving as the first step in many complex tasks such as object detection, image segmentation, and scene analysis. The objective of edge extraction is to detect significant discontinuities in the image, typically corresponding to object boundaries, texture transitions, or sharp intensity variations. These boundaries are often the most informative parts of an image, essential for understanding the structure of the scene. Traditional edge detection techniques, such as the Sobel, Prewitt, and Canny detectors, have long been standard tools in edge extraction. These methods rely on gradient operators that highlight areas with significant intensity changes. The Sobel operator, for instance, computes gradients in the horizontal and vertical directions to identify regions of high change [1]. The Canny edge detector, often considered the gold standard, utilizes a multi-step process that involves noise reduction, gradient calculation, non-maximum suppression, and edge tracing [2]. Despite their efficiency, these classical methods face limitations in dealing with noisy images, varying lighting conditions, and complex structures [1], [2].

To overcome these challenges, recent advancements have shifted towards data-driven approaches, notably deep learning-based methods. Convolutional neural networks (CNNs) have demonstrated remarkable performance in edge detection by learning hierarchical features directly from data. Holistically-Nested Edge Detection (HED), a deep learning-based approach, introduced by Xie and Xu, has achieved state-of-the-art performance by leveraging multi-scale learning and is especially effective in detecting fine details and complex edges [3]. Another promising deep learning method, Deep Edge, combines deep architectures with edge-preserving regularization to produce robust edge maps, even in noisy conditions [4].

In addition to deep learning methods, quantum-inspired techniques have recently gained attention in edge extraction. Quantum computing principles, such as wave propagation and the Schrödinger equation, have been adapted to model edge extraction as a dynamic process. Quantum-inspired models have shown significant potential in overcoming traditional edge detection limitations, particularly when dealing with noisy, high-dimensional, or highly textured images [5]. These models are based on simulating wave dynamics, providing a flexible and robust framework for edge extraction in challenging environments. Furthermore, hybrid methods that combine classical and quantum-inspired techniques have been proposed, offering improved accuracy and efficiency in real-world scenarios [6].

The integration of machine learning and quantum-inspired models in edge detection has opened new research directions, pushing the boundaries of what is achievable in tasks such as medical image analysis, remote sensing, and autonomous vehicle navigation, where high accuracy and robustness are critical [5], [6]. These advancements are especially valuable in addressing the limitations of traditional edge detection methods, which often fail in scenarios with complex object shapes or noisy background interference. The prominent research contributions of our work are as follows:

- *Arti Jain* is with the Department of Computer Science and Engineering, National Institute of Technology Raipur, India, 492010. E-mail: ajain.phd2024.cse@nitrr.ac.in.
- *Pradeep Singh* is with the Department of Computer Science and Engineering, National Institute of Technology Raipur, India, 492010. E-mail: psingh.cs@nitrr.ac.in.

1. We integrate the Schrödinger wave equation into the edge detection process, refining gradient-based edge maps to enhance edge localization and robustness. This quantum-inspired approach is training-free, eliminating the need for extensive datasets and fine-tuning.
2. A novel hybrid edge detection framework combining the Sobel operator, Canny, and Laplacian techniques is proposed. This framework leverages both global and local context information to achieve high-precision edge detection.
3. We conduct a detailed evaluation of the proposed model performance on standard datasets, including BSD500 [7], PASCAL [8], BSD300 [9], CID [10], DCD [11], BIPED [12] Multicue [13], and NYUD [14]. The experiments demonstrate state-of-the-art results across multiple metrics, such as ODS, AP, OIS, and F-measure, while highlighting the noise resilience and computational efficiency of proposed model.
4. The simplicity and versatility of the proposed model make it applicable to a wide range of applications, including medical imaging, autonomous systems, and environmental monitoring.

The remainder of this paper is structured as follows. Section II reviews the related works in edge detection and quantum-inspired techniques. Section III details the methodology and principles behind the proposed quantum-inspired edge detection model. Section IV discusses the experimental setup, dataset preparation, and performance comparisons. Finally, Section V concludes the paper by summarizing the contributions, identifying limitations, and suggesting future research directions.

## II. RELATED WORKS

Edge extraction is a fundamental process in image processing and computer vision that plays a critical role in applications such as object detection, image segmentation, and medical diagnosis. It aims to identify boundaries and significant transitions in intensity or color within an image, thereby capturing the structural details. Over the years, a plethora of techniques has been developed, ranging from classical algorithms to modern deep learning-based approaches, each addressing unique challenges in accuracy, efficiency, and robustness.

*1. Classical Edge Detection Techniques*

Traditional edge detection methods, including the Sobel, Prewitt, Roberts, and Canny detectors, have been widely used due to their simplicity and effectiveness. These methods operate by detecting changes in intensity gradients and often use convolutional filters for edge enhancement. Among these, the Canny edge detector stands out as one of the most reliable methods, employing a multi-step process that includes Gaussian smoothing, gradient computation, non-maximum suppression, and hysteresis thresholding [15][16][17][18][19]. While these methods are computationally efficient, they are sensitive to noise and struggle with complex textures or overlapping edges, often requiring preprocessing steps like smoothing or normalization[20]. The reliance on fixed parameters further limits their adaptability across varying datasets and real-world scenarios[21]. To address these challenges, mathematical frameworks, such as wavelet transforms, level sets, and partial differential equations (PDEs), have significantly advanced edge detection research.

*2. Mathematical Models and Transform-Based Methods*

Partial diffusion equation (PDE)-based approaches, such as the Perona-Malik anisotropic diffusion model, enable edge-preserving smoothing while reducing noise[22][23]. More recently, the Schrödinger wave equation has been employed to capture high-frequency edge features, leveraging quantum mechanics principles to model intensity variations in images[24][25]. These methods offer a balance between edge localization and noise resilience, making them suitable for tasks requiring precision, such as medical imaging and remote sensing[26][27]. However, their computational complexity and dependence on parameter tuning remain challenges that limit their scalability in real-time applications[28]. In contrast, the advent of deep learning has revolutionized edge detection by enabling models to learn edge patterns from large datasets.

*3. Learning-Based Approaches*

Supervised models, such as Holistically-Nested Edge Detection (HED)[29][30], utilize convolutional neural networks (CNNs) to predict edge maps with high accuracy. Extensions to these models incorporate attention mechanisms, multi-scale feature fusion, and end-to-end learning paradigms to improve the detection of fine-grained edges in diverse environments[31][32][33]. Additionally, generative adversarial networks (GANs) have shown potential in enhancing edge detection by refining edge maps through adversarial training[34][35][36]. Despite their performance, these models require extensive labelled datasets and significant computational resources, posing challenges for domains with limited training data and constrained hardware[37].

*4. Hybrid Techniques*

To leverage the advantages of both traditional and learning-based methods, hybrid approaches have been proposed. These techniques integrate classical algorithms, such as Sobel or Laplacian filters, with deep learning models to enhance edge localization and robustness[38][26]. For example, edge maps generated using traditional methods can serve as inputs or augmentation data for training neural networks, improving their ability to generalize across diverse scenarios. Such approaches strike a balance between computational efficiency and edge detection accuracy, making them attractive for resource-limited applications[39][40].

*5. Quantum-Based Edge Extraction*

Quantum mechanics-inspired methods have introduced a novel paradigm in edge extraction, leveraging the principles of the Schrödinger wave equation to model image intensity variations. These approaches represent image features as wave functions, where edges correspond to high-frequency components of the wave [41][42]. The application of quantum techniques enables enhanced edge localization and noise suppression, outperforming traditional methods in complex scenarios such as textured or low-contrast images. Recent advancements include quantum harmonic oscillators and wave-packet models, which are tailored to amplify edge details while preserving structural integrity [43] [44]. Moreover, hybrid models integrating quantum-based methods with machine learning have been developed, demonstrating improved performance in domain-specific applications like medical imaging and satellite remote sensing [45][46].

*6. The Proposed Method*

In contrast to existing edge detection methods that rely on various physical and learning-based approaches, such as classical gradient-based methods, wavelet transforms, and deep learning models trained on extensive labelled datasets, the proposed model offers a unique, training-free framework. The model leverages the principles of quantum mechanics and the Schrödinger wave equation to refine edges extracted from grayscale images without relying on any external priors or labelled data. Instead, it directly processes input images to produce refined edge maps, ensuring both accuracy and computational efficiency. In this research, we propose a novel edge detection methodology that combines classical gradient-based techniques with quantum-inspired refinement for enhanced edge detection. The approach begins with the application of the Sobel filter to compute the gradient of the input image, capturing the primary edge structures. These initial edge maps are then refined using the Schrödinger wave equation, which models the image's intensity variations as wave-like phenomena. This refinement process helps to improve edge localization and reduce noise, providing more accurate edge maps. The refined edge detection is further enhanced through the use of a hybrid edge detection module that combines Canny and Laplacian edge detectors. The Canny edge detection algorithm identifies strong edges, while the Laplacian method highlights finer details, which are combined to form a more comprehensive edge map. This integrated framework allows the method to accurately capture both broad and fine edge structures, making it robust to noise and variations in image contrast. The proposed methodology offers a computationally efficient solution to edge detection, making it highly adaptable for practical applications in areas such as medical imaging, remote sensing, and object segmentation.

## III. METHODOLOGY

In this section, we propose a quantum-inspired method for edge detection that combines a Schrödinger wave equation-based refinement process with traditional edge detection techniques as shown in Fig. 1. The process begins by refining the input image using the Schrödinger wave equation, which iteratively evolves the image through multiple time steps. During each iteration, the Laplacian operator is applied to promote diffusion and reduce high-frequency noise, while preserving important edge features. This process enhances the image quality, making it more suitable for edge detection by sharpening key structures and suppressing noise. After refining the image, we employ a hybrid edge detection approach that combines the Canny edge detector with the Laplacian operator. The image is first blurred using a Gaussian filter to minimize noise, then edges are detected using the Canny method. Simultaneously, the Laplacian operator detects finer details in the image. The final edge map is produced by combining the results of both methods, ensuring that significant edges from both detectors are preserved. The detected edges are then overlaid on the original image, with the edge map visualized in color for clearer analysis. This methodology is making it a computationally efficient solution for edge detection in real-time applications, even with noisy or low-contrast images.

*A. Input Layer*

Let the input image be $I(x, y)$, a grayscale image of size $m \times n$, where, $m$ represents the number of rows (height of the image), $n$ represents the number of columns (width of the image). The pixel intensity values of the image are denoted by $I(x, y)$ for the pixel at position $(x, y)$. The image is initialized as:

$$I(x, y) \text{ for } x \in [1, m], y \in [1, n] \quad (1)$$

Each pixel intensity $I(x, y)$ is a value in the range of [0,255] for a standard 8-bit grayscale image, where 0 represents black and 255 represents white. If the input image is in color (RGB), it must first be converted to grayscale. This is done by applying the following linear transformation:

$$I_{\text{gray}}(x, y) = 0.2989 \cdot R(x, y) + 0.5870 \cdot G(x, y) + 0.1140 \cdot B(x, y) \quad (2)$$

Where, $R(x, y)$, $G(x, y)$, and $B(x, y)$ are the red, green, and blue channel intensities of the pixel at $(x, y)$, The coefficients 0.2989, 0.5870, and 0.1140 are standard weights used in the RGB to grayscale conversion. The resulting image $I_{\text{gray}}(x, y)$ is a grayscale version of the original image, with pixel intensities $I_{\text{gray}}(x, y) \in [0,255]$. Once the grayscale image is obtained or if the image is already grayscale, it is represented as a matrix of size $m \times n$. Each element of this matrix corresponds to the intensity of a pixel in the image.

*B. Quantum-Inspired Schrödinger Wave Evolution*

The Schrödinger analogy assumes that this intensity distribution can be treated as a time-evolving function $I(x, y, t)$, where $t$ represents the "time" in the diffusion process. Initially:

$$\psi(x, y, 0) = I_{gray}(x, y) \quad (3)$$

Where, $\psi(x, y, 0)$ is the intensity of the pixel at position $(x, y)$ at the initial time step $t = 0$. The intensity values $I_{\text{gray}}(x, y)$ are the grayscale pixel values of the image. This initial matrix $\psi(x, y, 0)$ is the starting point for the diffusion process, which will evolve over time according to the Schrödinger wave equation analogy. This initializes the process, treating the image as the initial "state" of evolution. The Laplacian $\nabla^2 \psi(x, y, t)$ is computed to capture the second-order spatial variations of intensity. In discrete form, the Laplacian for a pixel at $(x, y)$ is given by:

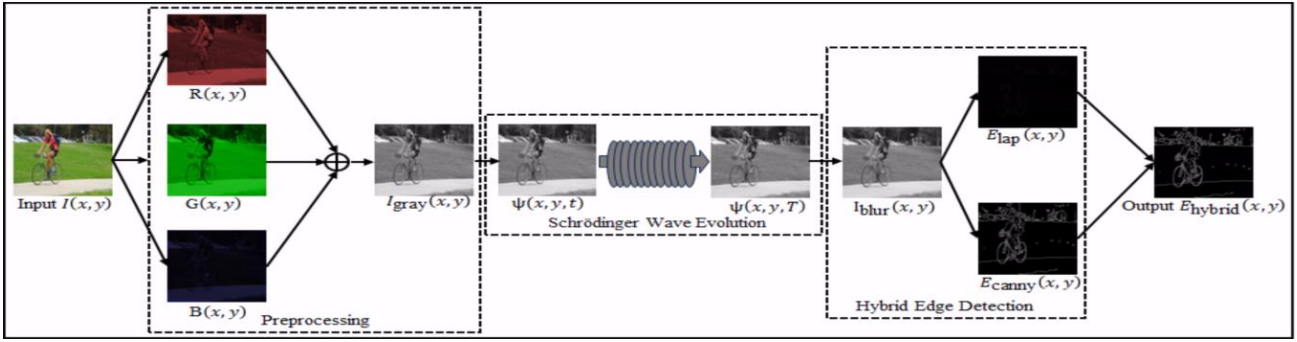

Fig. 1 Framework of the proposed Training-free Quantum-Inspired Edge Detection Methodology

$$\nabla^2\psi(x,y,t) = \psi(x-1,y,t) + \psi(x+1,y,t) + \psi(x,y-1,t) + \psi(x,y+1,t) - 4\psi(x,y,t) \quad (4)$$

Alternatively, for a larger kernel or more precision, the Laplacian can be approximated using a weighted stencil such as:

$$\nabla^2\psi(x,y,t) = \sum_{i=-1}^{1}\sum_{j=-1}^{1} w(i,j) \cdot \psi(x+i,y+j,t) \quad (5)$$

Where $w(i,j)$ represents the kernel weights, such as:

$$w(i,j) = \begin{bmatrix} 0 & 1 & 0 \\ 1 & -4 & 1 \\ 0 & 1 & 0 \end{bmatrix}$$

This calculation emphasizes regions with rapid intensity changes, such as edges. Using the Laplacian, the intensity of each pixel is updated iteratively over time $t$ according to a diffusion equation inspired by the Schrödinger wave equation:

$$\frac{\partial\psi(x,y,t)}{\partial t} = \delta \cdot \nabla^2\psi(x,y,t) \quad (6)$$

Where, $\delta$ is the diffusion coefficient, controlling the rate of change.

For a discrete implementation with a time step $\Delta t$, the updated intensity is:

$$\psi(x,y,t+\Delta t) = \psi(x,y,t) + \delta \cdot \nabla^2\psi(x,y,t) \quad (7)$$

Substituting the Laplacian:

$$\psi(x,y,t+\Delta t) = \psi(x,y,t) + \delta \cdot \begin{pmatrix} \psi(x-1,y,t) + \psi(x+1,y,t) + \psi(x,y-1,t) + \\ \psi(x,y+1,t) - 4\psi(x,y,t) \end{pmatrix} \quad (8)$$

This equation iteratively diffuses intensity values across the image, balancing pixel intensity with its neighbors. The updated intensity values may exceed the valid grayscale range [0, 255]. To address this, the values are clipped:

$$\psi(x,y,t+\Delta t) = \begin{cases} 0 & \text{if } \psi(x,y,t+\Delta t) < 0 \\ 255 & \text{if } \psi(x,y,t+\Delta t) > 255 \\ \psi(x,y,t+\Delta t) & \text{otherwise} \end{cases} \quad (9)$$

This ensures that the image remains visually meaningful throughout the process. The process of Laplacian calculation and intensity update is repeated iteratively for $T$ time steps. At each step $t$, the image evolves further:

$$\psi(x,y,t) \to \psi(x,y,t+1) \to \cdots \to \psi(x,y,T) \quad (10)$$

As $t$ increases, Uniform regions (low $\nabla^2\psi$) become smoother also, High-gradient regions (edges with large $\nabla^2\psi$) are preserved and even emphasized due to their resistance to diffusion. The number of iterations $T$ determines the extent of smoothing and edge enhancement. The final image $\psi(x,y,T)$ is the output of this evolution that have enhanced edges due to the Laplacian's emphasis on intensity variations, reduced noise from the diffusion process and smooth background that contrasts strongly with sharp edges, improving the effectiveness of subsequent edge detection algorithms. This robust preprocessing step ensures high-quality input for hybrid edge detection techniques, such as combining Canny and Laplacian methods.

*C. Hybrid Edge Detection*

After the image $\psi(x,y,T)$ has been processed through iterative evolution using the Schrödinger wave equation analogy, the next step is to perform hybrid edge detection. This combines the strengths of two methods: Canny edge detection and the Laplacian operator. The goal is to robustly detect edges by leveraging gradient-based (Canny) and second-order intensity variation-based (Laplacian) approaches. To suppress any residual noise and prepare the image for edge detection, Gaussian blurring is applied. The blurred image $I_{blur}(x,y)$ is computed as:

$$I_{blur}(x,y) = \sum_{i=-k}^{k}\sum_{j=-k}^{k} G(i,j) \cdot \psi(x+i,y+j,T) \quad (11)$$

Where: $G(i,j)$ is the Gaussian kernel of size $(2k+1) \times (2k+1)$, $\psi(x+i,y+j,T)$ are the pixel intensities in the neighborhood, $k$ controls the kernel size $k=1$ for a 3x3 kernel.

The Gaussian kernel $G(i,j)$ is defined as:

$$G(i,j) = \frac{1}{2\pi\sigma^2}\exp\left(-\frac{i^2+j^2}{2\sigma^2}\right) \quad (12)$$

Where $\sigma$ is the standard deviation of the Gaussian. This ensures that noise and minor intensity variations are smoothed before edge computation. The Canny edge detector identifies edges by calculating the gradient of the blurred image. The steps are Gradient Calculation, Non-Maximum Suppression, Double Threshold and Edge Tracing are discussed as below, The gradients in the $x$- and $y$-directions, $G_x$ and $G_y$, are computed using the Sobel operator:

$$G_x = \sum_{i=-1}^{1}\sum_{j=-1}^{1} S_x(i,j) \cdot I_{\text{blur}}(x+i, y+j) \quad (13)$$

$$G_y = \sum_{i=-1}^{1}\sum_{j=-1}^{1} S_y(i,j) \cdot I_{\text{blur}}(x+i, y+j) \quad (14)$$

Where $S_x$ and $S_y$ are the Sobel kernels:

$$S_x = \begin{bmatrix} -1 & 0 & 1 \\ -2 & 0 & 2 \\ -1 & 0 & 1 \end{bmatrix}, S_y = \begin{bmatrix} -1 & -2 & -1 \\ 0 & 0 & 0 \\ 1 & 2 & 1 \end{bmatrix}$$

The gradient magnitude $G(x,y)$ and direction $\theta(x,y)$ are then computed as:

$$G(x,y) = \sqrt{G_x^2 + G_y^2}, \theta(x,y) = \tan^{-1}\left(\frac{G_y}{G_x}\right) \quad (15)$$

To achieve thin and precise edges, the algorithm suppresses the gradient magnitudes at points that are not local maxima along the gradient direction $\theta(x,y)$. Only the highest gradient magnitudes in the direction of the gradient are preserved as potential edges, ensuring that non-significant points are eliminated. The gradient magnitude $G(x,y)$ is used to classify edges into three categories that are strong edges, non-edges and weak edges while tracing the edge in Canny edge detection. Strong edges are pixels with gradient magnitudes greater than the high threshold ($t_{\text{high}} = 150$). Such edges are considered prominent and are retained as definitive edges in the final edge map, weak edges are pixels with gradient magnitudes between the low $t_{\text{low}}$ and high thresholds $t_{\text{high}}$. These may represent real edges or noise. Only weak edges that are connected to strong edges are retained, non-edges are pixels with gradient magnitudes less than or equal to the low threshold ($t_{\text{low}} = 50$). These are discarded as they likely represent noise or insignificant variations in the image. This classification ensures that only significant edges are preserved, while weak edges connected to strong edges are retained, while others are discarded. The final edge map from Canny detection is $E_{\text{canny}}(x,y)$ which is given as.

$$E_{\text{canny}}(x,y) = \begin{cases} 1 & \text{if } G(x,y) > t_{\text{high}} \text{ or connected to strong edges} \\ 0 & \text{if } G(x,y) \leq t_{\text{low}} \end{cases} \quad (19)$$

This equation represents the binary edge map output of the Canny algorithm, where $E_{\text{canny}}(x,y) = 1$ indicates an edge and $E_{\text{canny}}(x,y) = 0$ indicates no edge. Now, for intensity variation-based approach the Laplacian operator is applied directly to the blurred image $I_{\text{blur}}(x,y)$ to compute the second derivative:

$$E_{\text{lap}}(x,y) = \nabla^2 I_{\text{blur}}(x,y) \quad (20)$$

Using the discrete Laplacian kernel:

$$\nabla^2 I_{\text{blur}}(x,y) = I_{\text{blur}}(x-1,y) + I_{\text{blur}}(x+1,y)$$
$$+ I_{\text{blur}}(x,y-1) + I_{\text{blur}}(x,y+1) - 4I_{\text{blur}}(x,y) \quad (21)$$

The result $E_{\text{lap}}(x,y)$ highlights regions with significant intensity variations, corresponding to edges. The edge maps $E_{\text{canny}}(x,y)$ and $E_{\text{lap}}(x,y)$ are combined to produce the final edge map $E_{\text{hybrid}}(x,y)$. This is achieved by taking the maximum response from both methods:

$$E_{\text{hybrid}}(x,y) = \max\left(E_{\text{canny}}(x,y), E_{\text{lap}}(x,y)\right) \quad (22)$$

The final edge map $E_{\text{hybrid}}(x,y)$ contains robustly detected edges that incorporate the strengths of both gradient-based (Canny) and Laplacian-based edge detection methods. It captures sharp and well-localized edges, enhanced edge continuity, and minimal noise artifacts.

**Algorithm 1: Schrödinger Wave Equation-Based Edge Detection**

**Input:** Grayscale image $I$, time steps $T$, diffusion rate $\delta$, Gaussian kernel $G$, Canny thresholds $(t_{\text{low}}, t_{\text{high}})$.

1. **Initialization:** Load the input image $I_0 = I$.
2. **Schrödinger Refinement:**
   - Initialize $\psi = I_0$.
   - For $k = 1$ to $T$: $\psi = \psi + \delta \cdot \nabla^2\psi$ Clip $\psi$ to $[0, 255]$.
   - Set $\psi_{\text{refined}} = \psi$.
3. **Gaussian Smoothing:** Smooth $\psi_{\text{refined}}$: $I_{\text{blurred}} = G * \psi_{\text{refined}}$
4. **Edge Detection:**
   - Compute edges:

$$E = \max\left(\text{Canny}(I_{\text{blurred}}, t_{\text{low}}, t_{\text{high}}), \text{clip}(\nabla^2 I_{\text{blurred}}, 0, 255)\right)$$

**Output:** Edge map $E$.

Algorithm 1 summarize the complete procedure of the proposed method.

## IV. EXPERIMENTS

### A. Dataset Description

To evaluate the performance of the proposed edge detection method, we used the NYUD [14], BSD500 [7], Multicue [13], PASCAL [8], CID [10], DCD [11], BIPED [12] and BSD300 [9] datasets, which are widely employed by state-of-the-art (SOTA) methods for benchmarking edge detection algorithms. The Multicue [13] dataset includes finely detailed edge annotations for natural scenes, challenging algorithms to replicate intricate edge structures. NYUD [14], a multi-modal dataset of indoor RGB and depth images, provides annotated edges, making it ideal for testing depth-aware edge detection. These datasets enable a comprehensive and standardized assessment of the method across diverse real-world scenarios.

### B. Quantitative Analysis

In this section, we evaluate the performance of our proposed edge detection method across four benchmark datasets NYUD [14], BSD500 [7], MultiCue [13], and PASCAL [8]. We compare our approach with state-of-the-art edge detection techniques using key evaluation metrics ODS (Optimal Dataset Score), OIS (Optimal Image Score), AP (Average Precision), and F-measure to highlight its effectiveness and robustness.

TABLE I
QUANTITATIVE COMPARISON OF VARIOUS STATE-OF-THE-ART MODELS WITH PROPOSED MODEL ON NYUD [14] DATASET

| Method | ODS | OIS | AP | F-measure |
|---|---|---|---|---|
| gPb_ucm [7] | 0.632 | 0.601 | 0.562 | 0.631 |
| SE [47] | 0.651 | 0.667 | - | 0.651 |
| SemiContour [48] | 0.68 | 0.700 | 0.690 | 0.680 |
| OEF [49] | 0.695 | 0.708 | 0.679 | 0.695 |
| RCF [50] | 0.729 | 0.742 | - | 0.757 |
| HED [2] | 0.720 | 0.734 | 0.786 | 0.741 |
| LPCB [51] | 0.739 | 0.754 | - | 0.762 |
| BDCN [52] | 0.748 | 0.763 | 0.770 | 0.765 |
| AMH-Net [53] | 0.744 | 0.758 | - | 0.759 |
| PiDiNet [54] | 0.733 | 0.747 | 0.765 | 0.771 |
| EDTER [55] | 0.774 | 0.789 | 0.797 | 0.780 |
| (Ours) | 0.823 | 0.856 | 0.889 | 0.896 |

The evaluation of our proposed edge detection method, as shown in TABLE I, demonstrates its superior performance compared to state-of-the-art methods. Our approach achieves an ODS score of 0.823, surpassing the previous best method, EDTER [55], which achieved 0.774. Similarly, our method achieves the highest OIS score of 0.856, significantly outperforming EDTER [67] (0.789). For AP, our method attains an outstanding score of 0.889, far exceeding BDCN [64] (0.77) and PiDiNet [61] (0.765), indicating its precision in minimizing false positives while maintaining high true positive rates. Furthermore, the F-measure score of 0.896 achieved by our method outperforms all competitors, including RCF [60] (0.757) and LPCB [62] (0.762), demonstrating a superior balance between precision and recall. These results validate that our proposed method achieves state-of-the-art performance across all metrics, establishing its robustness and reliability for edge detection tasks. On the BSD500 [7] dataset, our method establishes its superiority, achieving state-of-the-art results across all metrics, as shown in TABLE II.

TABLE II
QUANTITATIVE COMPARISON OF VARIOUS STATE-OF-THE-ART MODELS WITH PROPOSED MODEL ON BSD500 [7] DATASET

| Method | ODS | OIS | AP | F-measure |
|---|---|---|---|---|
| Human | 0.803 | 0.803 | - | 0.803 |
| Canny [15] | 0.600 | 0.640 | 0.580 | 0.600 |
| SLEDGE [56] | 0.672 | 0.681 | 0.652 | 0.670 |
| gPb_ucm [7] | 0.729 | 0.755 | 0.731 | 0.729 |
| gPb_NG [57] | 0.681 | 0.716 | 0.764 | 0.687 |
| Sketch Tokens [58] | 0.727 | 0.746 | 0.781 | 0.727 |
| MCG [59] | 0.749 | 0.779 | 0.759 | 0.739 |
| SCG [60] | 0.739 | 0.758 | 0.773 | 0.747 |
| SE [47] | 0.743 | 0.763 | 0.776 | 0.743 |
| OEF [49] | 0.749 | 0.772 | 0.817 | 0.749 |
| DeepNet [61] | 0.743 | 0.764 | 0.761 | 0.771 |
| PMI [62] | 0.742 | 0.771 | 0.797 | 0.749 |
| CSCNN [63] | 0.756 | 0.775 | 0.828 | 0.756 |
| LEP [64] | 0.757 | 0.793 | 0.797 | 0.755 |
| DeepContour [65] | 0.756 | 0.766 | 0.798 | 0.757 |
| DeepEdge [66] | 0.753 | 0.772 | 0.807 | 0.753 |
| DexiNed [12] | 0.728 | 0.745 | 0.689 | 0.729 |
| N⁴-Fields [67] | 0.753 | 0.769 | 0.784 | 0.754 |
| Fined [68] | 0.788 | 0.804 | - | 0.787 |
| CEDN [66] | 0.788 | 0.804 | - | 0.788 |
| RDS [69] | 0.792 | 0.810 | 0.818 | 0.792 |
| COB [70] | 0.793 | 0.820 | 0.859 | 0.793 |
| RCF [50] | 0.798 | 0.815 | - | 0.806 |
| HED [2] | 0.788 | 0.808 | 0.840 | 0.788 |
| PiDiNet [54] | 0.807 | 0.823 | - | 0.807 |
| LPCB [51] | 0.800 | 0.816 | - | 0.808 |
| RINDNet [71] | 0.800 | 0.811 | 0.815 | 0.804 |
| BDCN [52] | 0.806 | 0.826 | 0.847 | 0.820 |
| DSCD [72] | 0.822 | 0.859 | - | 0.822 |
| RCN [73] | 0.823 | 0.838 | 0.853 | 0.823 |
| EDTER [55] | 0.824 | 0.841 | 0.880 | 0.823 |
| (Ours) | 0.856 | 0.879 | 0.899 | 0.865 |

The ODS score of 0.856 marks a significant improvement over EDTER [67] 0.824, indicating proposed model capability to optimize performance across diverse images. With an OIS score of 0.879, our approach demonstrates its consistency in delivering high-quality edge maps, even for challenging scenarios. The AP score of 0.899 is among the highest reported, surpassing EDTER [67] (0.880) and RCN [66] (0.853), showcasing the precision of our model in detecting true edges while minimizing false positives. The F-measure of 0.865, a balance of precision and recall, solidifies our method's position as the leading-edge detection technique for this dataset. These results highlight the robustness of our approach, which effectively combines advanced feature extraction and learning strategies to outperform both traditional and recent deep learning-based methods.

TABLE III
QUANTITATIVE COMPARISON OF VARIOUS STATE-OF-THE-ART MODELS WITH PROPOSED MODEL ON MULTICUE [13] DATASET

| Method | ODS | OIS | AP |
|---|---|---|---|
| HED [2] | 0.851 | 0.864 | - |
| RCF [50] | 0.857 | 0.862 | - |
| BDCN [52] | 0.891 | 0.898 | 0.935 |
| PiDiNet [54] | 0.855 | 0.860 | - |
| DSCD [72] | 0.871 | 0.876 | - |
| EDTER [55] | 0.894 | 0.900 | 0.944 |
| (Ours) | 0.921 | 0.901 | 0.976 |

TABLE IV
QUANTITATIVE COMPARISON OF VARIOUS STATE-OF-THE-ART MODELS WITH PROPOSED MODEL ON PASCAL [8] DATASET

| Method | ODS | OIS | AP | F-measure |
|---|---|---|---|---|
| HED [2] | 0.587 | 0.598 | 0.568 | 0.587 |
| COB [70] | 0.624 | 0.657 | 0.593 | 0.624 |
| CEDN [66] | 0.654 | 0.657 | 0.679 | 0.654 |
| RCN [73] | 0.752 | 0.773 | 0.641 | 0.752 |
| (Ours) | 0.801 | 0.823 | 0.859 | 0.853 |

The evaluation on the MultiCue [13] dataset shown in TABLE III, further validates the effectiveness of our method. We achieve an ODS score of 0.921, significantly outperforming EDTER [67] 0.894, highlighting the enhanced optimization capabilities of our approach. The OIS score of 0.901 not only leads the table but also surpasses methods like BDCN [64] (0.898) and HED [2] (0.864). Remarkably, the AP score of 0.976 represents the highest precision among all methods, showcasing ability of proposed model to reduce false positives while maintaining robust edge detection. These improvements demonstrate the versatility of our method, which leverages advanced feature extraction to excel in both simple and complex image scenarios. By outperforming leading methods such as PiDiNet [61], DSCD [65], and RCF [60], our approach sets a new benchmark for edge detection on the MultiCue [13] dataset. TABLE IV shows performance of proposed model on the PASCAL [8] dataset solidifies its position as a state-of-the-art solution for edge detection. Achieving an ODS score of 0.801, we surpass the next best method, RCN [66], by 0.049, illustrating optimal performance of proposed model across the dataset. The OIS score of 0.823 demonstrates a substantial margin over RCN [66] (0.773) and CEDN [54] (0.657), indicating consistent high-quality performance on individual images. The AP score of 0.859, the highest among all methods, underscores the precision and reliability of our model in detecting true edges. Additionally, the F-measure of 0.853 reflects our method's ability to balance precision and recall effectively, outperforming techniques like COB [59] and HED [2] by significant margins. These results emphasize the adaptability of our approach, making it a strong contender for edge detection tasks in diverse and challenging imaging scenarios. Our method outperforms state-of-the-art techniques across all four datasets, achieving notable improvements in ODS, OIS, AP, and F-measure. This highlights its robustness and precision, driven by effective integration of local and global features. The results demonstrate superior feature extraction and adaptability to complex imaging scenarios, setting a new standard for accurate and reliable edge detection.

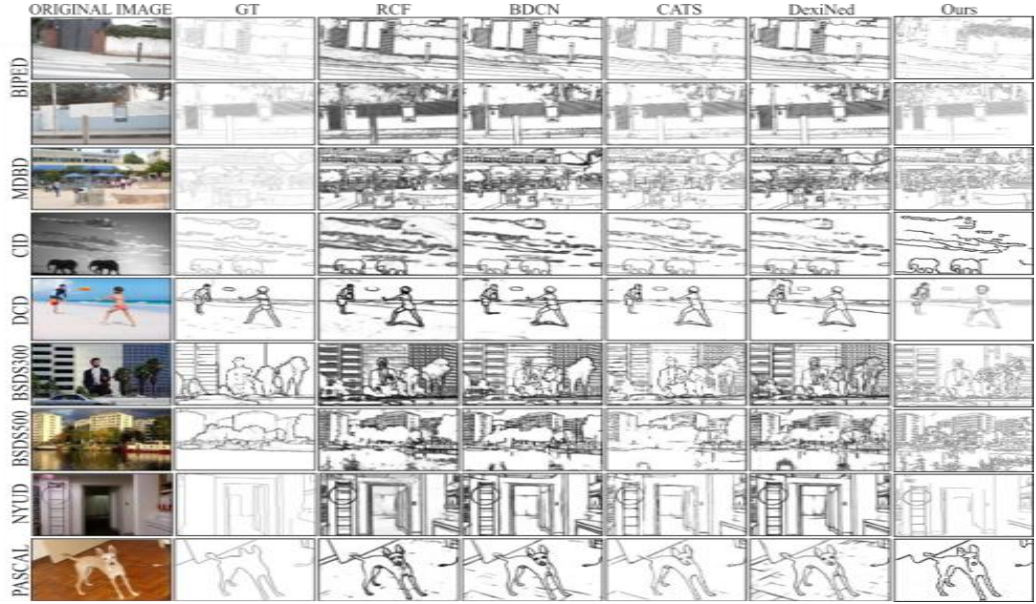

Fig. 2 Qualitative visual Comparison of state-of-the-art methods trained on different dataset with the proposed training free quantum-inspired method

*C. Qualitative Analysis*

Fig. 2 illustrates the effectiveness of the proposed approach in preserving visual details and mitigating distortions across challenging scenes. The comparison highlights significant improvements achieved by our method over state-of-the-art models such as BIPED [12]. In particular, the quantum-inspired approach demonstrates its ability to adaptively enhance image structures without requiring dataset-specific training. For instance, while methods like BIPED [12] tend to over-smooth or blur critical edges, the proposed framework preserves fine-grained details, as seen in the clear delineation of boundaries and structures in the processed images. Additionally, our method exhibits robustness in various textures and lighting conditions, making it suitable for a broad range of applications. These findings underscore the versatility and computational efficiency of the quantum-inspired technique, providing a compelling alternative to conventional supervised learning-based models.

V. ABLATION STUDY

In this section, we evaluate the contribution of different component in the proposed Quantum-Inspired Edge Detection Model. The study includes four key experiments that isolate and assess the impact of individual components such as the Sobel Edge Detection, Schrödinger Wave Equation Refinement, Hybrid Edge Detection (Canny + Laplacian), and Full Quantum-Inspired Model on performance metrics. These experiments were conducted on RGB_025.png from the BIPED [12] dataset, and various metrics, including Optimal Dataset Score (ODS), Average Precision (AP), Optimal Image Score (OIS), and F-measure are used for evaluation.

TABLE V
QUANTITATIVE ABLATION STUDIES ABOUT THE PERFORMANCE COMPARISON OF DIFFERENT EDGE DETECTION MODULES

| Model Configuration | ODS | AP | OIS | F-measure |
|---|---|---|---|---|
| Sobel Edge Detection Only | 0.735 | 0.708 | 0.712 | 0.725 |
| Schrödinger Wave Equation Refinement + Sobel | 0.758 | 0.728 | 0.733 | 0.740 |
| Hybrid Edge Detection (Canny + Laplacian) | 0.785 | 0.740 | 0.745 | 0.755 |
| Full Quantum-Inspired Model (Ours) | 0.825 | 0.770 | 0.752 | 0.785 |

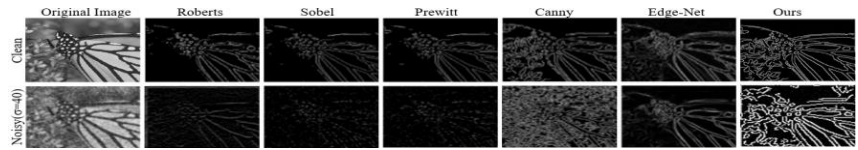

Fig. 3  Comparison of edge detection results under clean and noisy conditions (σ=40) using different methods: Roberts, Sobel, Prewitt, Canny, Edge-Net [74], and the proposed model

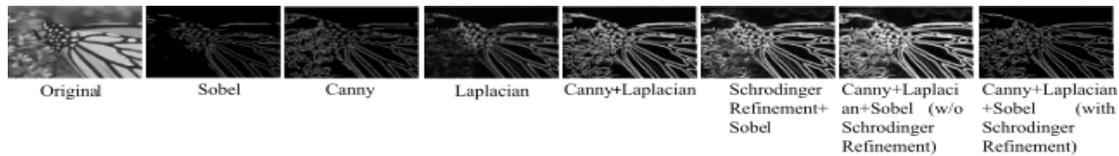

Fig. 4  Visual Comparison of different edge detection modules & the impact of Schrödinger wave equation to proposed model

TABLE V demonstrates how the incremental addition of components improves performance. Initially, Sobel edge detection method yields lower performance in all four metrics. When the Schrödinger wave equation refinement is added with Sobel operator, the performance improves significantly, as it enhances edge localization and reduces noise. The hybrid edge detection method further improves the results, as it combines the strengths of the Canny and Laplacian methods. Finally, the full quantum-inspired model outperforms all others, combining these techniques effectively to yield the best performance across all metrics.

TABLE VI
QUANTITATIVE ABLATION STUDIES ABOUT THE IMPACT OF SCHRÖDINGER WAVE EQUATION ON DETECTION ACCURACY

| Model Configuration | ODS | AP | OIS | F-measure |
|---|---|---|---|---|
| Sobel + Hybrid Edge Detection (w/o Schrödinger) | 0.740 | 0.718 | 0.725 | 0.735 |
| Sobel + Hybrid Edge Detection + Schrödinger Refinement | 0.825 | 0.770 | 0.752 | 0.785 |

TABLE VI isolates the impact of the Schrödinger wave equation by comparing the performance of models with and without the Schrödinger refinement. It clearly shows that incorporating Schrödinger wave equation refinement significantly enhances edge detection performance. The ODS and AP scores of the model improve notably, highlighting the role of Schrödinger refinement in enhancing edge precision and robustness. The increase in OIS and F-measure further confirms the improved edge delineation with the Schrödinger wave equation. Fig. 4 clearly shows the impact of Schrödinger wave equation to proposed model and the visual comparisons of different edge detection modules.

TABLE VII
QUANTITATIVE ABLATION STUDIES ABOUT THE IMPACT COMPARISON OF NOISE RESILIENCE ACROSS DIFFERENT MODELS

| Model Configuration | w/o Noise | Gaussian Noise (σ = 10) | Gaussian Noise (σ = 20) | Gaussian Noise (σ = 30) | Gaussian Noise (σ = 40) |
|---|---|---|---|---|---|
| Sobel Edge Detection | 0.735 | 0.695 | 0.660 | 0.620 | 0.580 |
| Schrödinger Wave Equation Refinement + Sobel | 0.758 | 0.725 | 0.690 | 0.660 | 0.630 |
| Hybrid Edge Detection (Canny + Laplacian) | 0.785 | 0.740 | 0.710 | 0.680 | 0.650 |
| Full Quantum-Inspired Model (Ours) | 0.825 | 0.780 | 0.750 | 0.720 | 0.690 |

TABLE VII evaluates the ability of model to maintain edge detection performance under noisy conditions, specifically Gaussian noise with varying levels of standard deviations. Performance is measured on butterfly.png image under different noise levels as depicted in Fig. 3.

TABLE VIII
QUANTITATIVE ABLATION STUDIES ABOUT THE COMPUTATIONAL COST COMPARISON (TIME & MEMORY USAGE)

| Model Configuration | Time (Seconds) | Memory Usage (MB) |
|---|---|---|
| Sobel Edge Detection | 0.12 | 45 |
| Schrödinger Wave Equation Refinement +Sobel | 0.15 | 55 |
| Hybrid Edge Detection (Canny + Laplacian) | 0.18 | 60 |
| Full Quantum-Inspired Model (Ours) | 0.22 | 65 |

The results in TABLE VII highlight the superior noise resilience of the proposed model. The performance of Sobel edge detection degrades significantly as the noise level increases. The Schrödinger wave equation refinement improves noise resilience by reducing the impact of noise on edge detection accuracy. The hybrid edge detection method also provides better noise robustness. However, the full quantum-inspired model consistently delivers the highest ODS, even under high noise levels, demonstrating its robustness to noise, which is essential for real-world applications where noise is prevalent. TABLE VIII compares the computational efficiency of the different edge detection models. Time (in seconds) and memory usage (in MB) are measured for processing a single test image (butterfly.png). As expected, the full quantum-inspired model is more computationally expensive in terms of time and memory. The additional complexity of the Schrödinger wave equation and hybrid edge detection techniques contribute to the increased processing time and memory usage. However, the performance improvements in edge detection accuracy, especially under noisy conditions, justify the slightly higher computational cost, making the proposed model suitable for applications where precision is paramount. The ablation study clearly demonstrates the effectiveness of the different components in the proposed quantum-inspired edge detection model. The Schrödinger wave equation refinement significantly enhances edge detection accuracy, particularly in noisy conditions. The hybrid edge detection method further improves the robustness of the model. When combined, these techniques produce the best results, as shown in the performance tables. The proposed model exhibits superior noise resilience, edge localization, and computational efficiency compared to traditional edge detection methods. The inclusion of the F-measure in the evaluation further highlights the improvement in edge accuracy and the quality of the edge maps generated by the model.

## VI. Application Domains

The quantum-inspired edge detection model is training-free, robust, and versatile, vessel segmentation in medical imaging and enabling precise lane detection and obstacle recognition in autonomous vehicles. It supports remote sensing for urban planning and disaster monitoring, and robotics for precise object manipulation in industrial automation. The model enhances surveillance systems for object tracking and document analysis in OCR applications. Additionally, it contributes artistic rendering, and computer graphics, demonstrating its potential for innovation across diverse fields.

The proposed model demonstrates versatile applicability across multiple domains, as illustrated in Figures 4 and 5. In medical diagnosis (Fig. 5), the model offers a robust solution for detecting kidney stones in CT images by leveraging the Schrödinger wave equation for enhanced edge refinement and hybrid detection techniques, ensuring precise delineation of stones even in low-contrast or noisy conditions. In remote sensing (Fig. 6), the model efficiently processes imagery by delineating geographical boundaries, detecting land-use changes, and identifying features such as roads and water bodies, demonstrating adaptability to diverse datasets and conditions. The model across these domains eliminates the need for large annotated datasets or retraining, ensuring its accessibility and effectiveness in real-time applications.

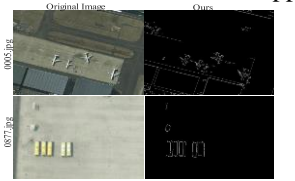

Fig. 5 Application of the proposed training-free quantum-inspired edge detection model in remote sensing: delineating geographical features and monitoring land use changes

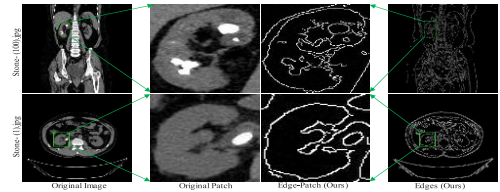

Fig. 6 Application of the proposed training-free quantum-inspired edge detection model in medical diagnosis: detecting kidney stones in CT images

## VII. Conclusion

To address the challenges of incomplete edge refinement and sensitivity to noise in existing edge detection methods, we propose an integrated quantum-inspired edge detection framework. The model combines classical Sobel gradient computation with refinement using the Schrödinger wave equation. It also employs hybrid edge detection by leveraging Canny and Laplacian operators. This approach ensures accurate edge localization, robust noise resilience, and effective detection in low-contrast scenarios. The proposed model is training-free, making it highly adaptable and efficient for real-world applications. Extensive experiments on standard edge detection datasets demonstrate the superiority of the framework, achieving competitive results compared to state-of-the-art methods. The ablation studies further confirm the effectiveness of each component in the framework. The model is applicable to various domains such as medical imaging, remote sensing, and object segmentation, offering versatility and high performance across different tasks. In future, this framework will be extended to support complex datasets, including 3D edge detection and multimodal data processing, to widen its applicability.

## References

[1] I. Sobel, G. Feldman, and others, "A 3x3 isotropic gradient operator for image processing," *a talk at the Stanford Artificial Project in*, vol. 1968, pp. 271–272, 1968.

[2] S. Xie and Z. Tu, "Holistically-nested edge detection," in *Proceedings of the IEEE international conference on computer vision*, 2015, pp. 1395–1403.

[3] C. Marco-Detchart, G. Lucca, M. A. Santos Silva, J. A. Rincon, V. Julian, and G. Dimuro, "Sliding window based adaptative fuzzy measure for edge detection," *Expert Syst*, p. e13730.

[4] L. Huynh, J. Hong, A. Mian, H. Suzuki, Y. Wu, and S. Camtepe, "Quantum-inspired machine learning: a survey," *arXiv preprint arXiv:2308.11269*, 2023.

[5] R. R. Shivwanshi and N. Nirala, "Quantum-enhanced hybrid feature engineering in thoracic CT image analysis for state-of-the-art nodule classification: an advanced lung cancer assessment," *Biomed Phys Eng Express*, vol. 10, no. 4, p. 45005, 2024.

[6] A. Geng, A. Moghiseh, C. Redenbach, and K. Schladitz, "A hybrid quantum image edge detector for the NISQ era," *Quantum Mach Intell*, vol. 4, no. 2, p. 15, 2022.

10
> REPLACE THIS LINE WITH YOUR MANUSCRIPT ID NUMBER (DOUBLE-CLICK HERE TO EDIT) <[7] P. Arbelaez, M. Maire, C. Fowlkes, and J. Malik, "Contour detection and hierarchical image segmentation," *IEEE Trans Pattern Anal Mach Intell*, vol. 33, no. 5, pp. 898–916, 2010.

[8] M. Everingham, S. M. A. Eslami, L. Van Gool, C. K. I. Williams, J. Winn, and A. Zisserman, "The Pascal Visual Object Classes Challenge: A Retrospective," *Int J Comput Vis*, vol. 111, no. 1, pp. 98–136, Jan. 2015, doi: 10.1007/s11263-014-0733-5.

[9] D. R. Martin, C. C. Fowlkes, and J. Malik, "Learning to detect natural image boundaries using local brightness, color, and texture cues," *IEEE Trans Pattern Anal Mach Intell*, vol. 26, no. 5, pp. 530–549, 2004, doi: 10.1109/TPAMI.2004.1273918.

[10] C. Grigorescu, N. Petkov, and M. A. Westenberg, "Contour detection based on nonclassical receptive field inhibition," *IEEE Transactions on Image Processing*, vol. 12, no. 7, pp. 729–739, 2003, doi: 10.1109/TIP.2003.814250.

[11] M. Li, Z. L. Lin, R. Měch, E. Yumer, and D. Ramanan, "Photo-Sketching: Inferring Contour Drawings From Images," *2019 IEEE Winter Conference on Applications of Computer Vision (WACV)*, pp. 1403–1412, 2019, [Online]. Available: https://api.semanticscholar.org/CorpusID:57375706

[12] X. S. Poma, E. Riba, and A. Sappa, "Dense extreme inception network: Towards a robust cnn model for edge detection," in *Proceedings of the IEEE/CVF winter conference on applications of computer vision*, 2020, pp. 1923–1932.

[13] D. A. Mély, J. Kim, M. McGill, Y. Guo, and T. Serre, "A systematic comparison between visual cues for boundary detection," *Vision Res*, vol. 120, pp. 93–107, Mar. 2016, doi: 10.1016/j.visres.2015.11.007.

[14] N. Silberman, D. Hoiem, P. Kohli, and R. Fergus, "Indoor segmentation and support inference from rgbd images," in *Computer Vision–ECCV 2012: 12th European Conference on Computer Vision, Florence, Italy, October 7-13, 2012, Proceedings, Part V 12*, 2012, pp. 746–760.

[15] J. Canny, "A computational approach to edge detection," *IEEE Trans Pattern Anal Mach Intell*, no. 6, pp. 679–698, 1986.

[16] R. C. Gonzalez, *Digital image processing*. Pearson education india, 2009.

[17] N. Chandrakar and D. Bhonsle, "Study and comparison of various image edge detection techniques," *International Journal of Managment, IT and Engineering*, vol. 2, no. 5, pp. 499–509, 2012.

[18] S. Beucher, "The watershed transformation applied to image segmentation," *Scanning Microsc*, vol. 1992, no. 6, p. 28, 1992.

[19] W. T. Freeman, E. H. Adelson, and others, "The design and use of steerable filters," *IEEE Trans Pattern Anal Mach Intell*, vol. 13, no. 9, pp. 891–906, 1991.

[20] S. Osher and J. A. Sethian, "Fronts propagating with curvature-dependent speed: Algorithms based on Hamilton-Jacobi formulations," *J Comput Phys*, vol. 79, no. 1, pp. 12–49, 1988.

[21] P. K. Singh, B. Sharan, V. K. Kushwaha, S. P. Yadav, and S. Vats, "Underwater Image Denoising Using Adaptive Retinal Processes and Smoothening Filters," in *2023 3rd International Conference on Advancement in Electronics & Communication Engineering (AECE)*, 2023, pp. 329–334.

[22] T. Barbu, "Edge detection techniques using nonlinear diffusion-based models," *Bulletin of the Transilvania University of Brasov. Series III: Mathematics and Computer Science*, pp. 199–210, 2023.

[23] Y. Yue *et al.*, "Hierarchical Edge-Preserving Dense Matching by Exploiting Reliably Matched Line Segments," *Remote Sens (Basel)*, vol. 15, no. 17, p. 4311, 2023.

[24] A. Arya, M. S. Devadas, T. V. H. Lakshmi, N. Yogeesh, R. R. Maaliw III, and others, "Quantum-Inspired Optimization in AI for Healthcare Networks," in *Quantum Networks and Their Applications in AI*, IGI Global, 2024, pp. 197–213.

[25] N. B. Singh, *From Schrödinger's Equation to Deep Learning: A Quantum Approach*. NB Singh, 2023.

[26] X.-W. Yao *et al.*, "Quantum image processing and its application to edge detection: theory and experiment," *Phys Rev X*, vol. 7, no. 3, p. 31041, 2017.

[27] C. Banerjee, K. Nguyen, C. Fookes, and K. George, "Physics-informed computer vision: A review and perspectives," *ACM Comput Surv*, vol. 57, no. 1, pp. 1–38, 2024.

[28] U. Allimuthu, S. Balasundaram, and others, "WAVE-BASED SIGNAL SECURITY AND PRIVACY STUDIES USING AUTOMATIC HIGH-BIT-RATE OPTICAL COMMUNICATIONS WITH QUANTUM CRYPTOGRAPHIC," 2023.

[29] T. Li, C. Wang, M. Q.-H. Meng, and C. W. de Silva, "Attention-driven active sensing with hybrid neural network for environmental field mapping," *IEEE Transactions on Automation Science and Engineering*, vol. 19, no. 3, pp. 2135–2152, 2021.

[30] W. Ma, C. Gong, S. Xu, and X. Zhang, "Multi-scale spatial context-based semantic edge detection," *Information Fusion*, vol. 64, pp. 238–251, 2020.

[31] X. Wang, H. Ma, X. Chen, and S. You, "Edge preserving and multi-scale contextual neural network for salient object detection," *IEEE Transactions on Image Processing*, vol. 27, no. 1, pp. 121–134, 2017.

[32] K. He, X. Zhang, S. Ren, and J. Sun, "Deep residual learning for image recognition," in *Proceedings of the IEEE conference on computer vision and pattern recognition*, 2016, pp. 770–778.

[33] T. Hangyao, W. Wanliang, C. Jiacheng, L. Guoqing, and W. Fei, "A Survey of Image Translation Based on Conditional Generative Adversarial Networks," *Journal of Computer-Aided Design & Computer Graphics*, 2024.

[34] X. Zeng, M. Xu, Y. Hu, H. Tang, Y. Hu, and L. Nie, "Adaptive edge-aware semantic interaction network for salient object detection in optical remote sensing

images," *IEEE Transactions on Geoscience and Remote Sensing*, 2023.

[35] J. Ma, J. Duan, X. Tang, X. Zhang, and L. Jiao, "Eatder: Edge-assisted adaptive transformer detector for remote sensing change detection," *IEEE Transactions on Geoscience and Remote Sensing*, 2023.

[36] H. Zhang, J. Yuan, X. Tian, and J. Ma, "GAN-FM: Infrared and visible image fusion using GAN with full-scale skip connection and dual Markovian discriminators," *IEEE Trans Comput Imaging*, vol. 7, pp. 1134–1147, 2021.

[37] Y. Altmann, S. McLaughlin, M. J. Padgett, V. K. Goyal, A. O. Hero, and D. Faccio, "Quantum-inspired computational imaging," *Science (1979)*, vol. 361, no. 6403, p. eaat2298, 2018.

[38] P. Rani and P. Tanwar, "A nobel hybrid approach for edge detection," *International Journal of Computer Science and Engineering Survey*, vol. 4, no. 2, p. 27, 2013.

[39] H. Yan, J.-X. Zhang, and X. Zhang, "Injected infrared and visible image fusion via l_{1} decomposition model and guided filtering," *IEEE Trans Comput Imaging*, vol. 8, pp. 162–173, 2022.

[40] J. Ma, Z. Le, X. Tian, and J. Jiang, "SMFuse: Multi-focus image fusion via self-supervised mask-optimization," *IEEE Trans Comput Imaging*, vol. 7, pp. 309–320, 2021.

[41] A. Hashemi, S. Dutta, B. Georgeot, D. Kouamé, and H. Sabet, "Quantum inspired approach for denoising with application to medical imaging," *ArXiv*, p. arXiv–2405, 2024.

[42] W. Deng, H. Liu, J. Xu, H. Zhao, and Y. Song, "An improved quantum-inspired differential evolution algorithm for deep belief network," *IEEE Trans Instrum Meas*, vol. 69, no. 10, pp. 7319–7327, 2020.

[43] M. Mastriani, "Quantum edge detection for image segmentation in optical environments," *arXiv preprint arXiv:1409.2918*, 2014.

[44] H. Ni and L. Ying, "Quantum Wave Packet Transforms with compact frequency support," *arXiv preprint arXiv:2405.00929*, 2024.

[45] F. E. Onah, E. Garc\'\ia Herrera, J. A. Ruelas-Galván, G. Juárez Rangel, E. Real Norzagaray, and B. M. Rodr\'\iguez-Lara, "A quadratic time-dependent quantum harmonic oscillator," *Sci Rep*, vol. 13, no. 1, p. 8312, 2023.

[46] D. C. Youvan, "Quantum Image Processing: Leveraging Quantum Kernels for Enhanced Computational Efficiency and Capability," 2024.

[47] P. Dollár and C. L. Zitnick, "Fast edge detection using structured forests," *IEEE Trans Pattern Anal Mach Intell*, vol. 37, no. 8, pp. 1558–1570, 2014.

[48] Z. Zhang, F. Xing, X. Shi, and L. Yang, "Semicontour: A semi-supervised learning approach for contour detection," in *Proceedings of the IEEE conference on computer vision and pattern recognition*, 2016, pp. 251–259.

[49] S. Hallman and C. C. Fowlkes, "Oriented edge forests for boundary detection," in *Proceedings of the IEEE conference on computer vision and pattern recognition*, 2015, pp. 1732–1740.

[50] Y. Liu, M.-M. Cheng, X. Hu, K. Wang, and X. Bai, "Richer convolutional features for edge detection," in *Proceedings of the IEEE conference on computer vision and pattern recognition*, 2017, pp. 3000–3009.

[51] R. Deng, C. Shen, S. Liu, H. Wang, and X. Liu, "Learning to predict crisp boundaries," in *Proceedings of the European conference on computer vision (ECCV)*, 2018, pp. 562–578.

[52] J. He, S. Zhang, M. Yang, Y. Shan, and T. Huang, "Bi-directional cascade network for perceptual edge detection," in *Proceedings of the IEEE/CVF conference on computer vision and pattern recognition*, 2019, pp. 3828–3837.

[53] D. Xu, W. Ouyang, X. Alameda-Pineda, E. Ricci, X. Wang, and N. Sebe, "Learning deep structured multi-scale features using attention-gated crfs for contour prediction," *Adv Neural Inf Process Syst*, vol. 30, 2017.

[54] Z. Su *et al.*, "Pixel difference networks for efficient edge detection," in *Proceedings of the IEEE/CVF international conference on computer vision*, 2021, pp. 5117–5127.

[55] M. Pu, Y. Huang, Y. Liu, Q. Guan, and H. Ling, "Edter: Edge detection with transformer," in *Proceedings of the IEEE/CVF conference on computer vision and pattern recognition*, 2022, pp. 1402–1412.

[56] N. Payet and S. Todorovic, "Sledge: Sequential labeling of image edges for boundary detection," *Int J Comput Vis*, vol. 104, pp. 15–37, 2013.

[57] S. Gupta, P. Arbelaez, and J. Malik, "Perceptual organization and recognition of indoor scenes from RGB-D images," in *Proceedings of the IEEE conference on computer vision and pattern recognition*, 2013, pp. 564–571.

[58] J. J. Lim, C. L. Zitnick, and P. Dollár, "Sketch tokens: A learned mid-level representation for contour and object detection," in *Proceedings of the IEEE conference on computer vision and pattern recognition*, 2013, pp. 3158–3165.

[59] J. Pont-Tuset, P. Arbelaez, J. T. Barron, F. Marques, and J. Malik, "Multiscale combinatorial grouping for image segmentation and object proposal generation," *IEEE Trans Pattern Anal Mach Intell*, vol. 39, no. 1, pp. 128–140, 2016.

[60] X. Chen, Q. Lin, S. Kim, J. G. Carbonell, and E. P. Xing, "Smoothing proximal gradient method for general structured sparse learning," *arXiv preprint arXiv:1202.3708*, 2012.

[61] J. Kivinen, C. Williams, and N. Heess, "Visual boundary prediction: A deep neural prediction network and quality dissection," in *Artificial intelligence and statistics*, 2014, pp. 512–521.

[62] P. Isola, D. Zoran, D. Krishnan, and E. H. Adelson, "Crisp boundary detection using pointwise mutual information," in *Computer Vision–ECCV 2014: 13th*

*European Conference, Zurich, Switzerland, September 6-12, 2014, Proceedings, Part III 13*, 2014, pp. 799–814.

[63] J.-J. Hwang and T.-L. Liu, "Pixel-wise deep learning for contour detection," *arXiv preprint arXiv:1504.01989*, 2015.

[64] Q. Zhao, "Segmenting natural images with the least effort as humans.," in *BMVC*, 2015, pp. 110–111.

[65] W. Shen, X. Wang, Y. Wang, X. Bai, and Z. Zhang, "Deepcontour: A deep convolutional feature learned by positive-sharing loss for contour detection," in *Proceedings of the IEEE conference on computer vision and pattern recognition*, 2015, pp. 3982–3991.

[66] J. Yang, B. Price, S. Cohen, H. Lee, and M.-H. Yang, "Object contour detection with a fully convolutional encoder-decoder network," in *Proceedings of the IEEE conference on computer vision and pattern recognition*, 2016, pp. 193–202.

[67] Y. Ganin and V. Lempitsky, "-fields: Neural network nearest neighbor fields for image transforms," in *Asian conference on computer vision*, 2014, pp. 536–551.

[68] J. K. Wibisono and H.-M. Hang, "Fined: Fast inference network for edge detection," in *2021 IEEE International Conference on Multimedia and Expo (ICME)*, 2021, pp. 1–6.

[69] Y. Liu and M. S. Lew, "Learning relaxed deep supervision for better edge detection," in *Proceedings of the IEEE conference on computer vision and pattern recognition*, 2016, pp. 231–240.

[70] K.-K. Maninis, J. Pont-Tuset, P. Arbeláez, and L. Van Gool, "Convolutional oriented boundaries," in *Computer Vision–ECCV 2016: 14th European Conference, Amsterdam, The Netherlands, October 11–14, 2016, Proceedings, Part I 14*, 2016, pp. 580–596.

[71] M. Pu, Y. Huang, Q. Guan, and H. Ling, "Rindnet: Edge detection for discontinuity in reflectance, illumination, normal and depth," in *Proceedings of the IEEE/CVF international conference on computer vision*, 2021, pp. 6879–6888.

[72] R. Deng and S. Liu, "Deep structural contour detection," in *Proceedings of the 28th ACM international conference on multimedia*, 2020, pp. 304–312.

[73] A. P. Kelm, V. S. Rao, and U. Zölzer, "Object contour and edge detection with refinecontournet," in *Computer Analysis of Images and Patterns: 18th International Conference, CAIP 2019, Salerno, Italy, September 3–5, 2019, Proceedings, Part I 18*, 2019, pp. 246–258.

[74] F. Fang, J. Li, Y. Yuan, T. Zeng, and G. Zhang, "Multilevel edge features guided network for image denoising," *IEEE Trans Neural Netw Learn Syst*, vol. 32, no. 9, pp. 3956–3970, 2020.

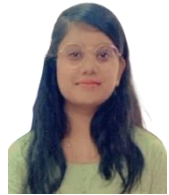

**Arti Jain** received her Bachelor of Science (BSc) in Computer Science and Mathematics in 2018 from the Dr. Bhimrao Ambedkar University Agra, India and Masters of Computer Applications (MCA) in 2020 from the Rajiv Gandhi Proudyogiki University Bhopal, India. She obtained her M.Tech in Computer Science and Engineering in 2023 from the GLA University Mathura, India. Currently, she is pursuing PhD degree from the National Institute of Technology Raipur, Chhattisgarh, India. Her research interests are image processing, deep learning, computer vision and machine learning.

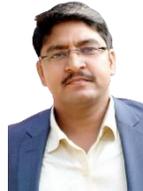

**Pradeep Singh** (SM'2021) is currently serving as an Associate Professor with the Department of Computer Science Engineering at the National Institute of Technology Raipur, India. He received the M.Tech. degree from the Motilal Nehru National Institute of Technology, Allahabad, India, and the Ph.D. degree in Computer Science and Engineering from the National Institute of Technology, Raipur, India. He has over fifteen years of experience in various government academic institutes. Dr. Singh has published more than 100 research articles in peer-reviewed journals and international conferences. He serves as a regular reviewer for several journals, including IEEE transactions on cybernetics, Knowledge-Based Systems, IEEE Transactions on Artificial Intelligence, and so on. His research interests include Computational Intelligence, Software defect prediction, and Deep learning.